\documentclass[5p,times]{elsarticle}
\usepackage{algorithm}
\usepackage{url}
\usepackage{graphicx}
\usepackage{subfigure}
\usepackage{color}
\usepackage{float}
\usepackage{sidecap}
\usepackage{rotating}
\usepackage{booktabs}
\usepackage{multirow}
\usepackage{ulem}
\usepackage{algpseudocode}
\usepackage{times}
\usepackage{amsmath}
\usepackage{amssymb}
\usepackage{commath}

\begin{document}
\begin{sloppypar}
\begin{frontmatter}

\title{Rotation invariant point cloud analysis: Where local geometry meets global topology}
\author[ad1]{Chen~Zhao}
\ead{hust\_zhao@hust.edu.cn}

\author[ad2]{Jiaqi~Yang}
\ead{jqyang@nwpu.edu.cn}

\author[ad1]{Xin~Xiong}
\ead{xiong\_xin@hust.edu.cn}

\author[ad1]{Angfan~Zhu}
\ead{zhuangfan@hust.edu.cn}

\author[ad1]{Zhiguo~Cao\corref{corre}}
\ead{zgcao@hust.edu.cn}
\cortext[corre]{Corresponding author.}

\author[ad3]{Xin~Li}
\ead{Xin.Li@mail.wvu.edu}

\address[ad1]{National Key Laboratory of Science and Technology on Multi-spectral Information Processing, School of Artificial Intelligence and Automation, Huazhong University of Science and Technology, Wuhan, 430074, China}
\address[ad2]{School of Computer Science, Northwestern Polytechnical University, Xi'an, 710072, China}
\address[ad3]{Lane Department of Computer Science and Electrical Engineering, West Virginia University, Morgantown, 26506, USA}

\begin{abstract}
Point cloud analysis is a fundamental task in 3D computer vision. Most previous works have conducted experiments on synthetic datasets with well-aligned data; while real-world point clouds are often not pre-aligned. How to achieve rotation invariance remains an open problem in point cloud analysis. To meet this challenge, we propose an approach toward achieving rotation-invariant (RI) representations by combining local geometry with global topology. In our local-global-representation (LGR)-Net, we have designed a two-branch network where one stream encodes local geometric RI features and the other encodes global topology-preserving RI features. Motivated by the observation that local geometry and global topology have different yet complementary RI responses in varying regions, two-branch RI features are fused by an innovative multi-layer perceptron (MLP) based attention module. To the best of our knowledge, this work is the first principled approach toward adaptively combining global and local information under the context of RI point cloud analysis. Extensive experiments have demonstrated that our LGR-Net achieves the state-of-the-art performance on various rotation-augmented versions of ModelNet40, ShapeNet, ScanObjectNN, and S3DIS. 
\end{abstract}


\end{frontmatter}


\section{Introduction}
\begin{figure}[t]
	\centering
	\subfigure[PointNet]
	{ \includegraphics[width=0.48\linewidth]{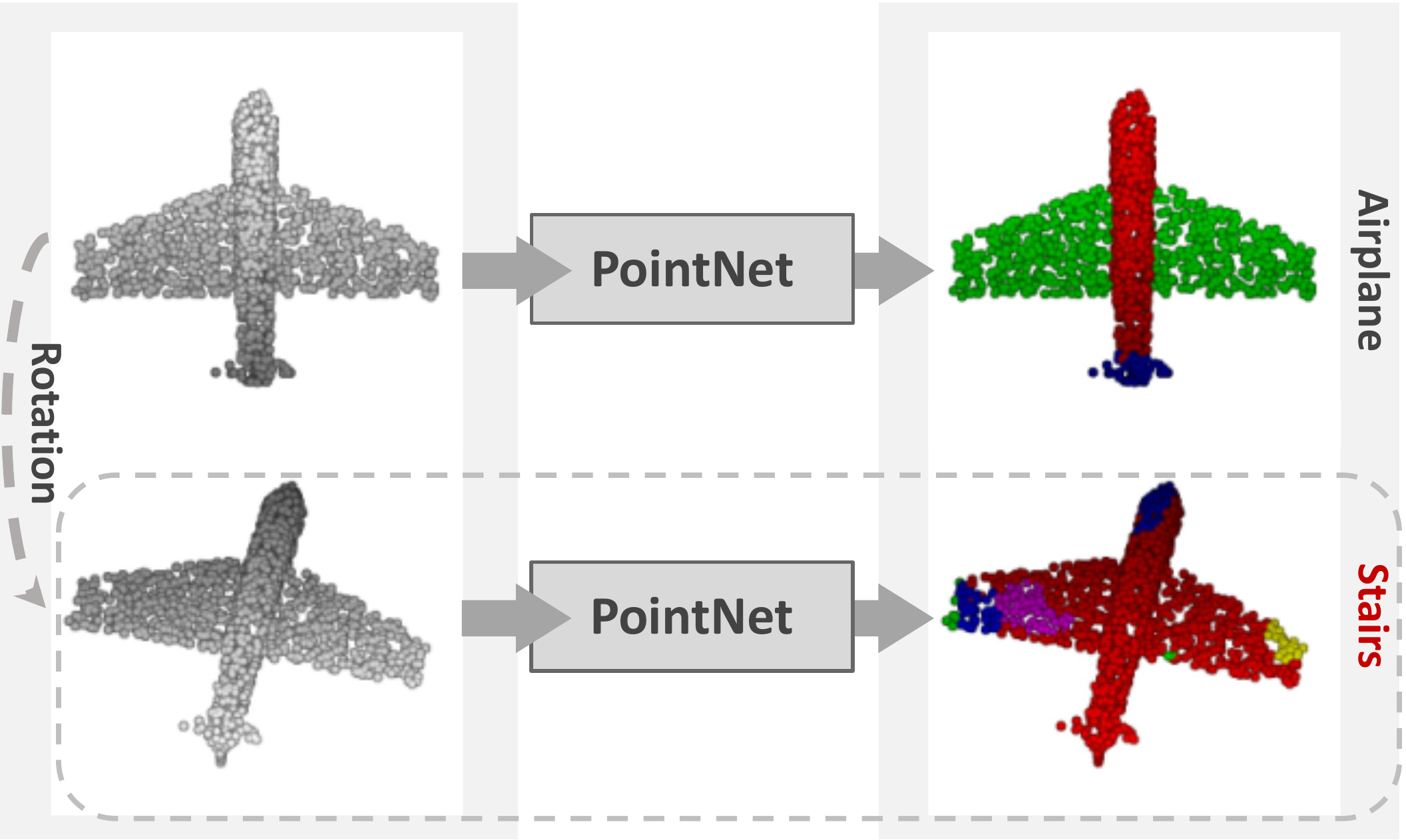}\label{fig:moti_a}
	}
	\subfigure[LGR-Net]
	{\includegraphics[width=0.48\linewidth]{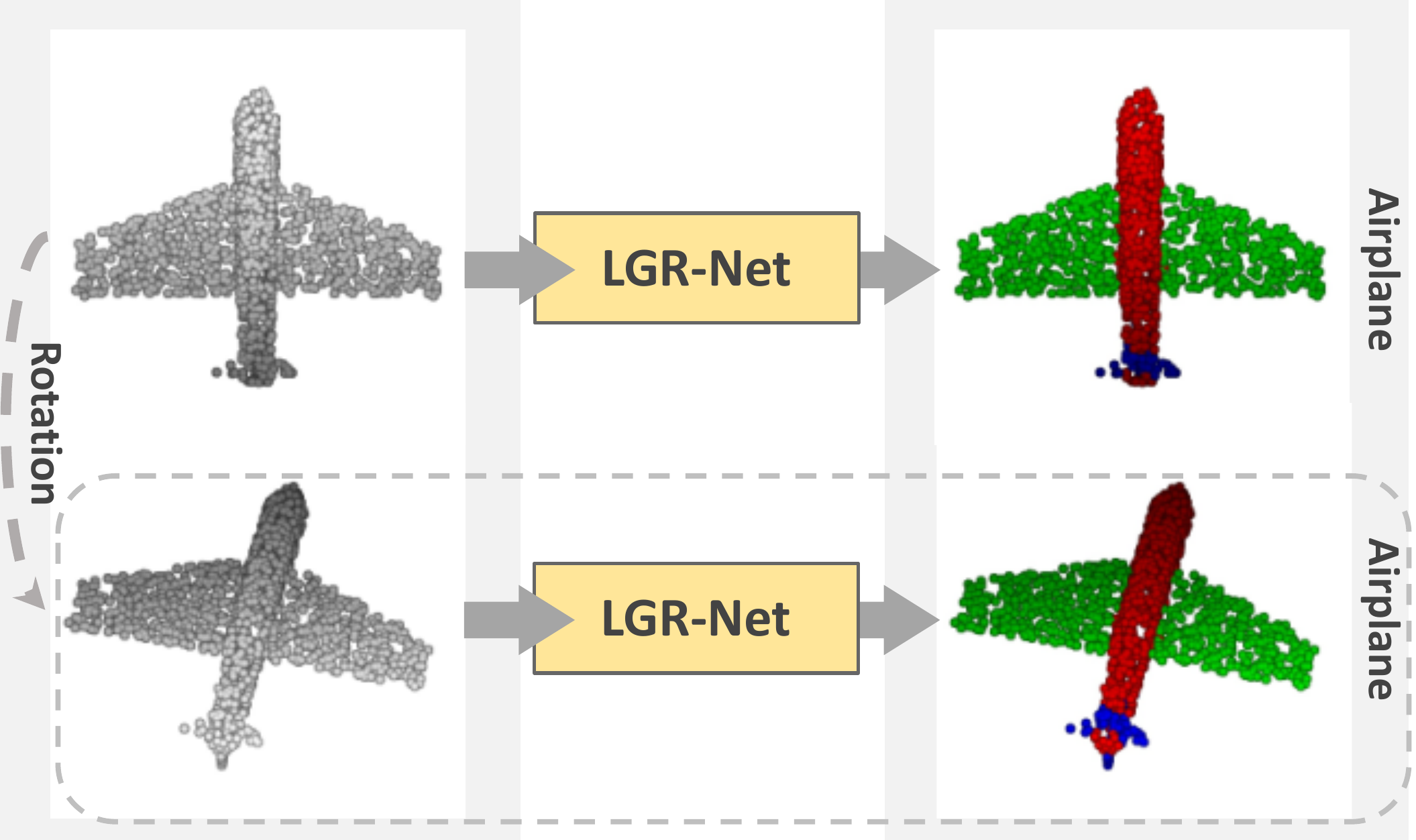}\label{fig:moti_b}
	}
	
	\caption{The proposed LGR-Net is rotation-invariant as compared to PointNet. The segmentation and classification results generated by PointNet (a) are considerably affected by the rotation; while the results of our LGR-Net (b) are invariant.}
\end{figure}

3D computer vision has been playing a pivotal role in many real-world applications -e.g., autonomous driving~\cite{li2019gs3d,meyer2019lasernet,yang2018ipod, cheng2016orthogonal}, augmented reality~\cite{alexiou2017towards, limberger2015real}, and robotics~\cite{durrant2006simultaneous,bailey2006simultaneous}. As a basic type of 3D data representation, point cloud analysis has received increasingly more attention in 3D vision. One of the pioneering works of deep learning based point cloud analysis, PointNet~\cite{qi2017pointnet}, employs multi-layer perceptron (MLP) to extract salient features from raw 3D coordinates. Unfortunately, most previous works are evaluated on synthetic datasets such as ModelNet40~\cite{wu20153d} and ShapeNet~\cite{yi2016scalable}, where point cloud models are assumed to be pre-aligned. Nonetheless, it is seldom the case to access well-aligned point clouds in real world applications, where geometric transformations are inevitable. In fact, the pose of point cloud models is often arbitrary in practice - it might include simple translation or complex 3D rotations or both. The performance of PointNet and its modified versions~\cite{qi2017pointnet++,xu2018spidercnn,wang2019dynamic} degrades rapidly due to the change of coordinates (caused by unknown geometric transformations). As shown in Fig.~\ref{fig:moti_a}, the accuracy of PointNet on both classification and segmentation significantly deteriorates in the presence of small rotations. 
\label{sec:int}

Note that translation invariance can be easily achieved by recentering point clouds. By contrast, rotation invariance is more challenging and has attracted increasingly more attention in recent years~\cite{zhang2019rotation,chen2019clusternet}. An intuitive solution to address the issue of rotation-invariant (RI) analysis is to augment the training data by considering all possible rotations. However, such ad-hoc strategy is practically infeasible because the space of all 3D rotations is too large to sample. Another sensible approach is to use Spherical Fourier Transform (SFT) \cite{wang2009rotational} on spheres to achieve rotation equivariance~\cite{esteves2018learning,cohen2018spherical}. Despite its theoretical appeal, discrete implementation of SFT is nontrivial (the loss of information is inevitable during the projection) and spherical CNNs often require extra processing (e.g., max pooling) to achieve rotation invariance. Alternatively, one can consider the pursuit of RI representations for point cloud analysis. The spatial coordinates of point clouds might change with rotations; but geometric attributes such as distances and angles do not vary. Indeed, RI representations have been considered most recently~\cite{zhang2019rotation,chen2019clusternet}. However, existing methods have primarily focused on utilizing geometric shape attributes in local regions only; while their distinctiveness is often questionable in the presence of symmetric structures (e.g., planes) (refer to Fig. \ref{fig:moti}). The potential of exploiting global topology-related RI representations has been largely overlooked as far as we know. 

In this paper, we present a simple yet effective solution to RI point cloud analysis by combining local geometry and global topology information in a principled manner. When compared against PointNet, our method dubbed LGR-Net is fully invariant to the rotation in both classification and segmentation tasks (as shown in Fig. \ref{fig:moti_b}). For local representations, we have extended persistent feature histograms~\cite{rusu2008aligning} into a more distinctive feature space, where the shape attributes in a local region around the query point are determined by a Darboux frame~\cite{spivak1970comprehensive}. For global representations, we generate RI spatial locations by projecting original points onto a rotation-equivariant global coordinate system established from the down-sampled skeleton-like structure~\cite{tagliasacchi2009curve,huang2013l1}. We both rigorously and experimentally show that our local and global representations (LGR) can achieve the desirable rotation invariance. In order to extract RI features from the LGR, we propose a two-branch network where the local and global information are separately processed and then adaptively combined by an attention-based fusion. The rationale behind the attention-based adaptive fusion can be justified by contrasting a point in a flat region with a point around a corner (refer to Fig. \ref{fig:moti}) - for the former, local information is clearly insufficient for classification or segmentation; while for the latter, local information is already amenable to describing the distinctive local structures. Extensive experimental results are reported for both synthetic datasets (ModelNet40~\cite{wu20153d} and ShapeNet~\cite{yi2016scalable}) and real-world datasets (ScanObjectNN~\cite{uy2019revisiting} and S3DIS~\cite{armeni20163d}) to show that our approach has achieved remarkably superior performance on rotation-augmented benchmark datasets.
In a nutshell, our major contributions are summarized as follows:
\begin{itemize}
    \item We present LGR-Net which considers local geometric features and global topology-preserving features to achieve rotation invariance. The complementary relationship between shape descriptions and spatial attributes is cleverly exploited by our two-branch network, and their strengths are adaptively combined by an attention-based fusion module.
	\item Our approach\footnote{Code is made available at \url{https://github.com/sailor-z/LGR-Net.}} achieves impressive experimental results when compared with current state-of-the-art methods on both synthetic and real-world datasets undergoing random 3D rotations. The performance improvements are particularly striking in the presence of complex rotations (e.g., $SO3$ group \cite{altmann2005rotations}).
\end{itemize}

\section{Related Work}
\label{sec:rel}
\noindent\textbf{Spatial transformations.}
In early works, a straightforward approach is to augment the training data using transformations of arbitrary rotations~\cite{qi2017pointnet,qi2017pointnet++}. However, since 3D rotations include three degrees of freedom -i.e., pitch, yaw, and roll, sampling them from $0$-degree to $360$-degree results in \emph{astronomical number of data points}. Consequently, it is often impractical to cover all possible rotations in real-world applications. An alternative yet more efficient approach employs deep learning methods to directly learn the unknown spatial transformations~\cite{qi2017pointnet}. Specifically, T-Net has been used in PointNet to regress a $3\times3$ spatial transformation and a $64\times64$ high-dimensional transformation, targeting at transforming point clouds into a canonical coordinate system. Nevertheless, PointNet with the learned transformations is still vulnerable to the nuisance of rotations as shown in Sec.~\ref{sec:exp}.
\\

\noindent\textbf{Rotation equivariance convolutions.}
Inspired by the popularity of convolutional neural networks in 2D computer vision, several works have been developed to leverage the success of convolutions from image data to point clouds~\cite{xu2018spidercnn,li2018pointcnn,wang2019dynamic}. However, most previous works did not take rotation invariance into account and therefore were sensitive to rotations. Some works have utilized spherical convolutions to achieve rotation equivariance~\cite{esteves2018learning,cohen2018spherical,liu2018deep}. 
Note that the equivariance means the output and the input vary equally, which is intrinsically different from invariance. Additional process such as max pooling becomes necessary in order to achieve rotation invariance. Moreover, the loss of information is inevitable during the generation of mesh/voxel, the transform and inverse transform, which limits the overall performance in practical implementations.
\\

\begin{figure}[!h]
	\centering
	\includegraphics[width=0.8\linewidth]{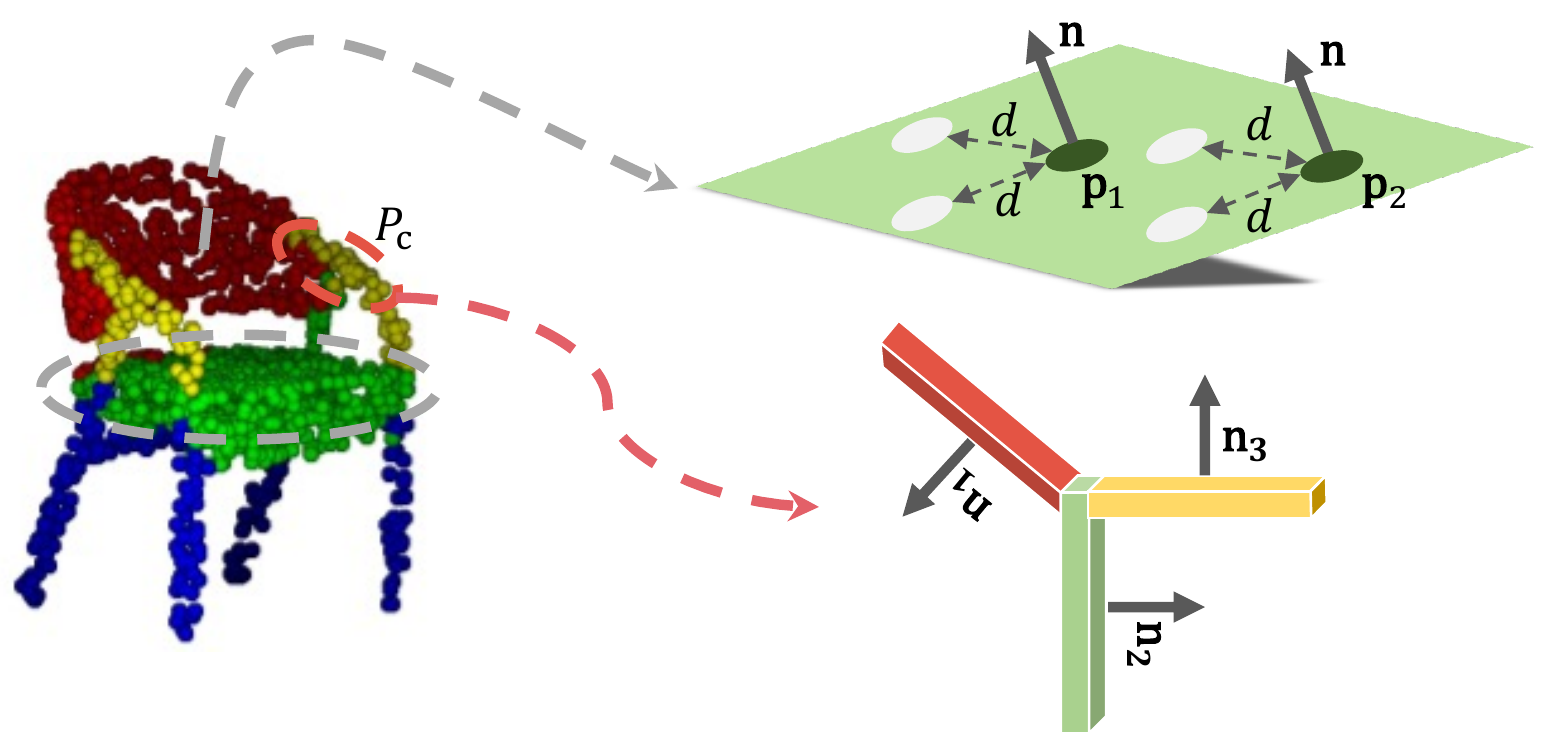}

	\caption{Effectiveness of local geometry in different regions. $(\mathbf{p}_1, \mathbf{p}_2)$ are located on the same plane with the normal $\mathbf{n}$, while $P_c$ is a set of corner points. $d$ represents the Euclidean distance. The same Euclidean distances and normals lead to ambiguous local descriptions around $\mathbf{p}_1$ and $\mathbf{p}_2$; while the local context around $P_c$ is distinctive.}
	\label{fig:moti}
\end{figure}

\noindent\textbf{Rotation-invariant representations.} The third class of approaches attempt to transform the raw point clouds into rotation-invariant representations. Among RI representations, distances and angles are the most widely used features. Specifically, a 4D point pair feature (PPF) was proposed in~\cite{deng2018ppf} for the task of RI descriptors, which utilized the distances and angles between the central reference point and neighbors in a local patch. For the tasks of classification and segmentation,~\cite{chen2019clusternet} integrated distance, angle, $sin$, and $cos$ features in local kNN-graphs into a cluster network;~\cite{zhang2019rotation} combined distance and angle features in local graphs and those in reference points generated by down sampling. We note that all previous works concentrated on local features -i.e., relative distances and angles in local graphs. However, local information is inevitably ambiguous for the tasks of classification and segmentation. For instance, the geometric shape descriptions represented by distances and angles tend to be similar among different points in the same plane (refer to Fig.~\ref{fig:moti}). Along this line of reasoning, absolute spatial attributes in a global coordinate system are critical to resolve location uncertainty.

\begin{figure*}[!h]
	\centering
	\includegraphics[width=0.9\linewidth]{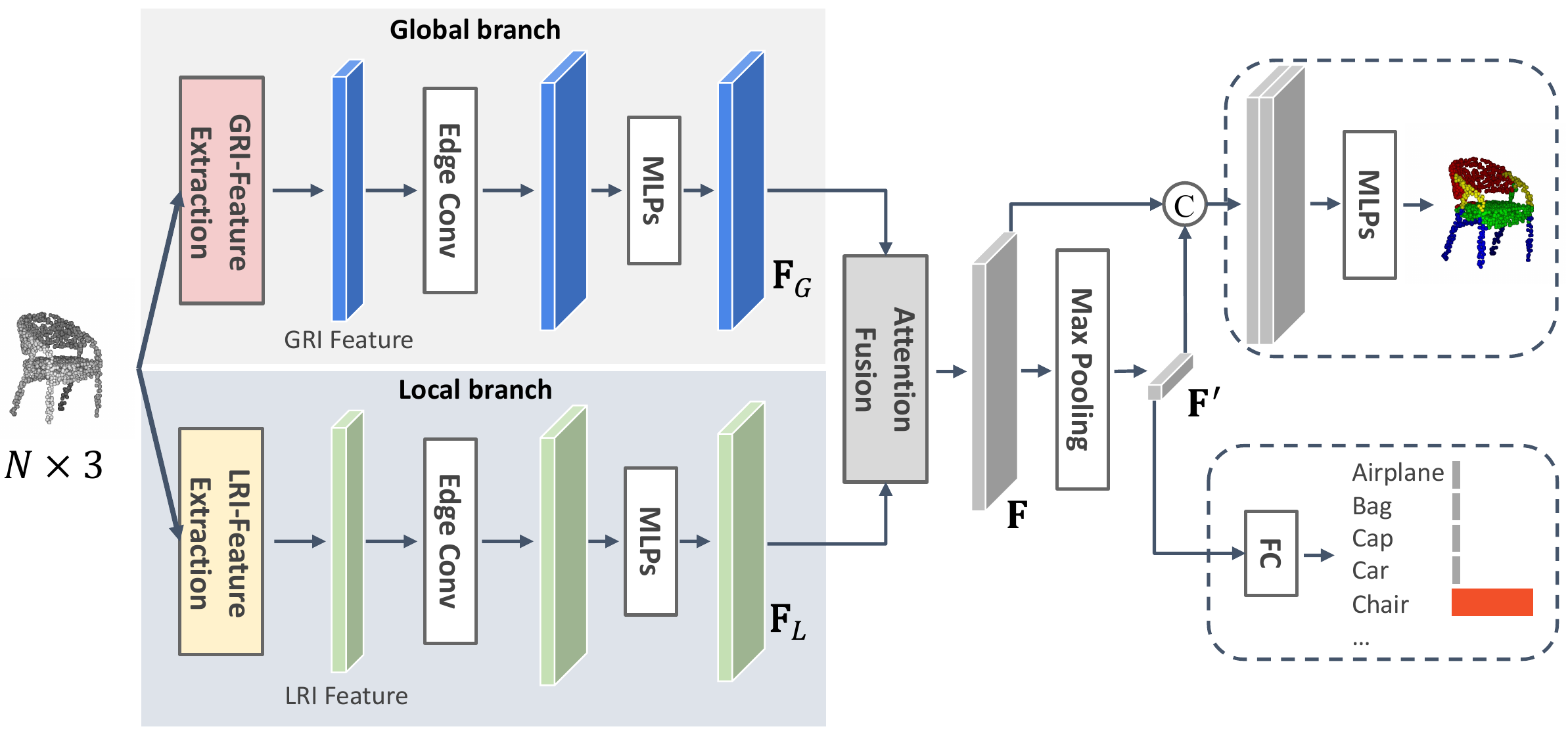}

	\caption{Network architecture. The architecture includes two branches that consume raw 3D points $\mathbf{P}\in\mathbb{R}^{N\times3}$ to separately generate local and global RI representations. High-dimensional features ($\mathbf{F}_G$, $\mathbf{F}_L$) are extracted by MLPs and then fused into a feature embedding ($\mathbf{F}$) by an attention-based fusion module. }
	\label{fig:network}
\end{figure*}

\section{Proposed Method}
Fig.~\ref{fig:network} shows the architecture of our LGR-Net. 3D points are consumed by a two-branch network to separately generate local and global RI representations. An attention-based fusion module is proposed to fuse the local and global features in an adjustable manner. The superiority of our separation-and-fusion design is introduced in Sec.~\ref{sec:abalation}.
\label{sec:met}

\subsection{Problem Statement}
Our method directly works with raw point cloud data, which are represented as a matrix of 3D coordinates $\mathbf{P}\in\mathbb{R}^{N\times3}$ - i.e., $\mathbf{P} = [\mathbf{p}_1; \mathbf{p}_2;\dots; \mathbf{p}_n]$ with $\mathbf{p}_i = (x_i, y_i, z_i)$. The normal of each point is denoted by $\mathbf{n}_i = (n^{i}_x, n^{i}_y, n^{i}_z)$. The issue of rotation invariance can be formulated by transforming $\mathbf{P}$ through a $3\times3$ orthogonal matrix $\mathbf{R}\in SO(3)$ ($det(\mathbf{R})=1$), which contains three degrees of freedom - i.e., $\alpha\in[0, 2\pi], \beta\in[0, \pi]$, and $\gamma\in[0, 2\pi]$. The objective of achieving rotation invariance then boils down to
\begin{equation}\label{eq:LRF1}
\mathcal{F}(\mathbf{P}\mathbf{R})=\mathcal{F}(\mathbf{P}),
\end{equation}
where $\mathcal{F}: \mathbb{R}^{N\times3}\rightarrow \mathbb{R}^{N\times D}$.
For the classification task, the desirable output is $s$ scores among which the maximum is expected to be the correct class label. For the segmentation task, the output is a $N\times m$ map whose entries indicate the scores of $m$ categories. Our objective is to achieve invariance to 3D rotations in both tasks.

\subsection{Local Branch}
\begin{figure}[!h]
	\centering
	\includegraphics[width=0.6\linewidth]{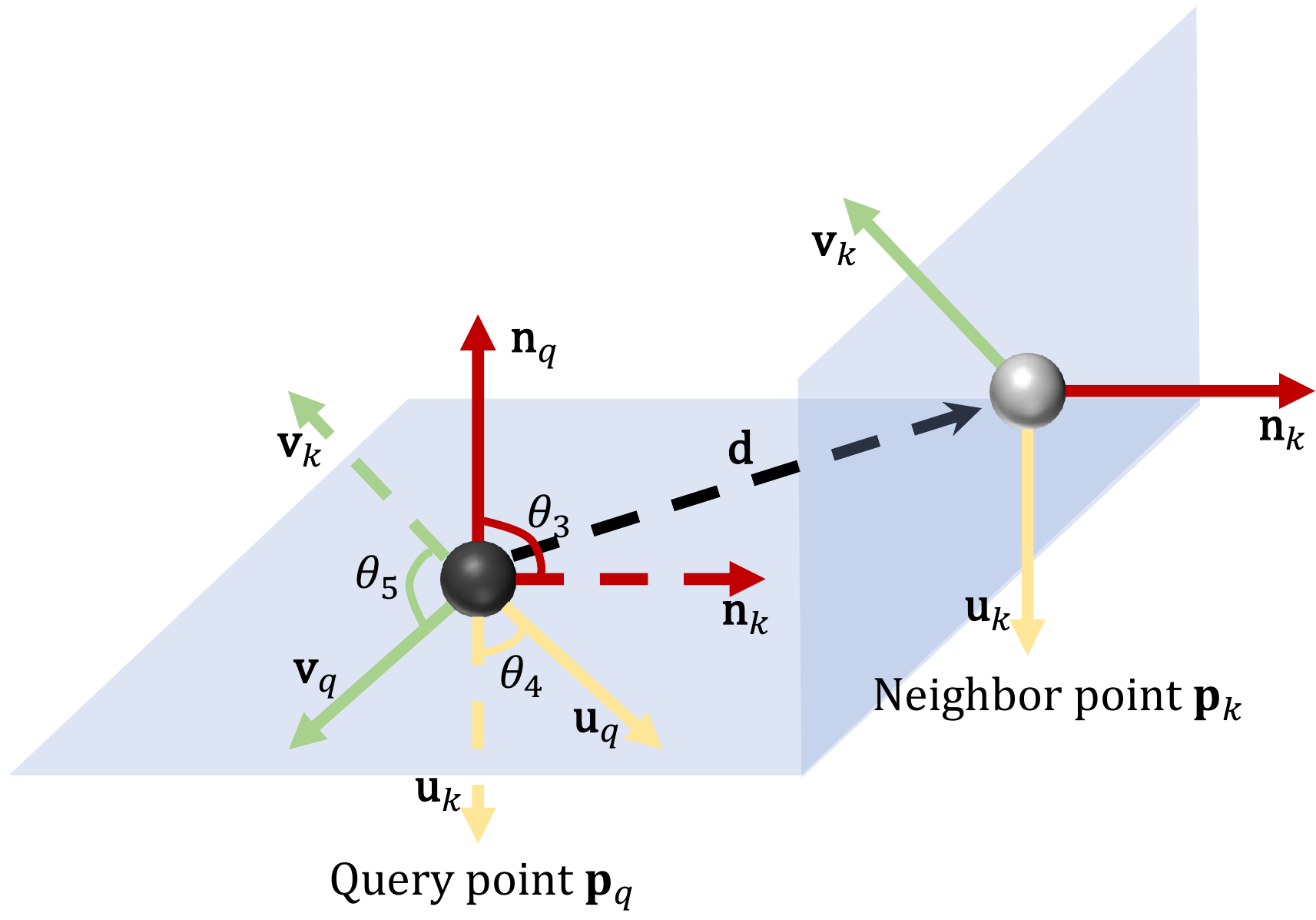}

	\caption{LRI (local-rotation-invariant)-feature extraction in the local branch. Given a query point $\mathbf{p}_q$ and its neighbors ($\mathbf{p}_k$ as an example), the local geometric shape is described by the relative distance $\left\|\mathbf{d}\right\|$ and high-order relationships between local coordinates $(\mathbf{n}_q, \mathbf{u}_q, \mathbf{v}_q)$ and $(\mathbf{n}_k, \mathbf{u}_k, \mathbf{v}_k)$, where $(\mathbf{n}_q,\mathbf{n}_k)$ are the normals of $(\mathbf{p}_q, \mathbf{p}_k)$, and other axes are located by cross product.}
	\label{fig:local}
\end{figure}


In previous works, local features have been proven critical to the tasks of point cloud classification and segmentation~\cite{xu2018spidercnn,li2018pointcnn,wang2019dynamic}, because local features are capable of describing geometric shape attributes in local regions. Taking rotation invariance into account, we employ the relative distances and angles as basic local descriptors, whose RI property can be easily verified. In order to have a richer collection of local shape descriptions, we consider an extended Darboux frame~\cite{rusu2008aligning,spivak1970comprehensive} by characterizing higher-order relationships among RI geometric attributes (distances and angles) as follows.

As shown in Fig.~\ref{fig:local}, For a query point $\mathbf{p}_q$, a local graph is generated by k-nearest neighbor (kNN) searching, and we assume a neighbor point $\mathbf{p}_k$ to be one of the kNNs. The relative position between $\mathbf{p}_q$ and $\mathbf{p}_k$ is described as $\left\|\mathbf{d}\right\|$ ($\mathbf{d} = \mathbf{p}_k-\mathbf{p}_q$). However, the location of $\mathbf{p}_k$ is ambiguous without taking orientation into account. Consequently, we estimate the orientation of $\mathbf{p}_k$ by calculating the higher-order relationships between the local coordinates centered at $\mathbf{p}_q$ and $\mathbf{p}_k$. 
Specifically, the local frame (e.g., $(\mathbf{n}_q, \mathbf{u}_q, \mathbf{v}_q)$) is generated as
\begin{equation}\label{eq:LRF2}
\mathbf{u}_q = \mathbf{d} \times \mathbf{n}_q,
\end{equation}
\begin{equation}\label{eq:LRF3}
\mathbf{v}_q = \mathbf{u}_q \times \mathbf{n}_q,
\end{equation}
where $\times$ denotes cross product. The septuplet $[\mathbf{d},(\mathbf{n}_q, \mathbf{u}_q, \mathbf{v}_q), \\(\mathbf{n}_k, \mathbf{u}_k, \mathbf{v}_k)]$ then serves as the building block for characterizing higher-order relative relationship. The orientation of $\mathbf{p}_k$ is computed by a 7-dimensional vector $\{\theta_1, \theta_2, \theta_3, \theta_4, \theta_5, \theta_6, \theta_7\}$, in which each entry denotes the angle between a pair of feature descriptors $(\mathbf{d}, \mathbf{n}_q)$, $(\mathbf{d}, \mathbf{n}_k)$, $(\mathbf{n}_q, \mathbf{n}_k)$, $(\mathbf{u}_q, \mathbf{u}_k)$, $(\mathbf{v}_q, \mathbf{v}_k)$, $(\mathbf{u}_q, \mathbf{v}_k)$, $(\mathbf{v}_q, \mathbf{u}_k)$, respectively. Moreover, each angle descriptor is defined by the $cos$ similarity between two low-order feature descriptors - taking $\theta_1$ as an example, we have
\begin{equation}\label{eq:LRF4}
cos(\theta_1) = \frac {\mathbf{d}\cdot\mathbf{n}_q}{\left\|\mathbf{d}\right\|\left\|\mathbf{n}_q\right\|}.
\end{equation}
Note that $\theta_6$ and $\theta_7$ are included to resolve the ambiguity arising from the orientation of local surfaces. Given $k$ nearest neighbors for a query point $\mathbf{p}_q$, a $k\times8$ RI representation is generated which comprehensively characterizes the local pairwise relative relationships around $\mathbf{p}_q$ while satisfies minimum redundancy requirement. 

It is worth noting that normals are employed in the local branch for local frame estimation instead of pure information enrichment. In previous works~\cite{qi2017pointnet, qi2017pointnet++}, normals are immediately concatenated with 3D coordinates to enrich the input information, which is vulnerable against rotations (Please refer to PointNet ($xyz+normal$) in Table~\ref{tab:modelnet}). By contrast, normals are reasonably utilized in our local branch to estimate local RI features which are not only capable of describing local geometric structures, but also invariant to rotations.
\subsection{Global Branch}
\begin{figure}[!h]
	\centering
	\includegraphics[width=0.9\linewidth]{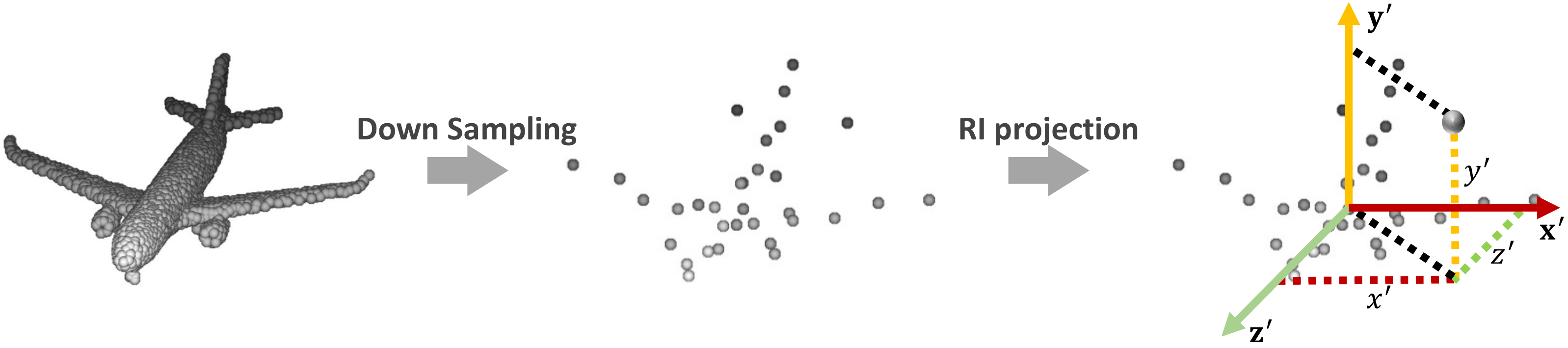}

	\caption{GRI (global-rotation-invariant)-feature extraction in the global branch. A rotation equivariance coordinate system is generated from a skeleton-like structure. Raw points are then projected onto the generated frame, which leads to rotation-invariant spatial locations.}
	\label{fig:global}
\end{figure}

Although geometric attributes in local regions have been studied in RI point cloud analysis, the issue of exploiting absolute spatial information has largely remained open. \cite{zhang2019rotation} suggests that the classification result significantly increases on rotation-free datasets when the presented (RI) representations are replaced with raw 3D coordinates, while the method is no longer robust to rotation. As shown in Fig.~\ref{fig:moti}, working with local feature descriptors alone is often insufficient due to the lack of distinctive spatial attributes such as absolute locations. Considering the points located on a plane, local geometric attributes (e.g., distances and angles) tend to be similar and cause inevitable confusion. Therefore, the rotation-invariant spatial attributes are expected to eliminate the ambiguity.

An intuitive solution to characterize spatial attributes is to work with point locations in a global coordinate system. However, the raw locations are sensitive to rotations. In order to acquire rotation-invariant spatial attributes, we employ singular value decomposition (SVD)~\cite{golub1971singular} which is a promising strategy capable of seeking canonical rotation-equivariant directions. Nevertheless, as shown in \cite{kazhdan2003rotation}, exploiting second-order shape information does not guarantee the optimal alignment. Moreover, it is noise-sensitive and time-consuming to directly apply SVD to the original point cloud model which may contain thousands of points. For the sake of an efficient and robust solution, we propose to down-sample the original point cloud, while preserving the global topological structure -i.e., skeleton-like structure as shown in Fig.~\ref{fig:global}. The actual down-sampling procedure is implemented by farthest point sampling \cite{moenning2003fast} in our experiments. The robustness of our down-sampled SVD strategy against nuisances can be found in Table~\ref{tab:scan} and Table~\ref{tab:rob}.
 
\begin{figure}[!h]
	\centering
	 \includegraphics[width=0.9\linewidth]{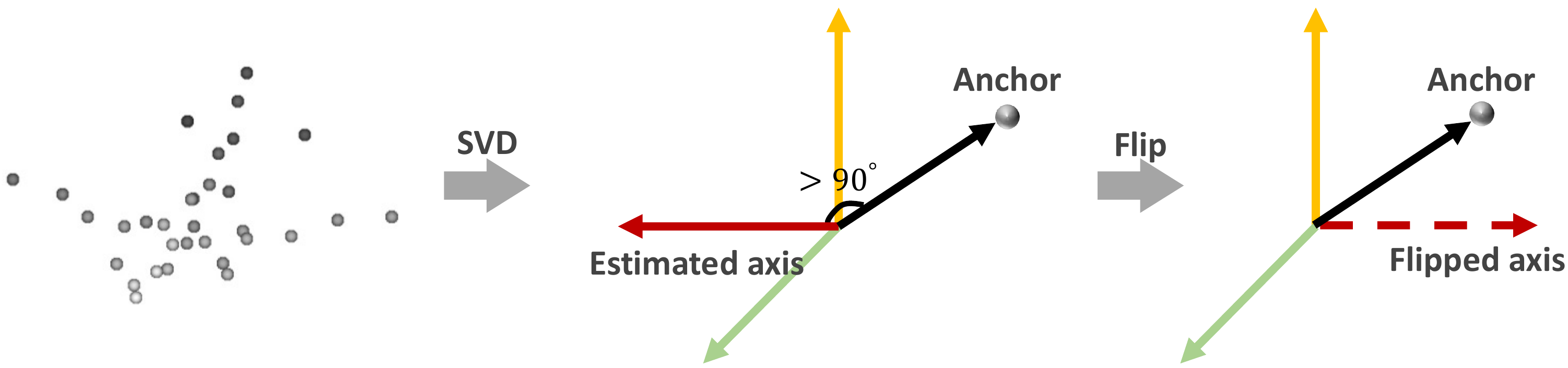}
	\caption{Illustration of determining the final directions of axes in the global branch.}
	\label{fig:flip}
\end{figure}

Then we carry out SVD on the down-sampled structure $\mathbf{P}_{d}$ which is formulated as
\begin{equation}\label{eq:LRF5}
\mathbf{U}\Sigma\mathbf{V}^{T} = \mathbf{P}_{d},
\end{equation}
where $\mathbf{V}$ contains the three orthogonal axes which are equivariant to rotations. To achieve rotation invariance, points are transformed from the original model to the newly-established global coordinate system as
\begin{equation}\label{eq:LRF6}
\hat{\mathbf{P}} = \mathbf{P}\cdot\mathbf{V},
\end{equation}
where $\hat{\mathbf{P}}$ describes the desired spatial RI attributes. Rigorous proof about rotation invariance of SVD-based transformation can be found in Sec.~\ref{sec:3.5}. Moreover, since the raw SVD technique is vulnerable to sign flipping, we address this issue by a simple yet effective solution. Specifically, we determine the final directions of three axes by estimating the angles between each axis and a predefined anchor point (the farthest point from the centroid in our experiments). As shown in Fig.~\ref{fig:flip}, the axis will be flipped if the corresponding angle is larger than $90^{\circ}$.
\subsection{Attention-based Fusion}

In order to generate an overall feature embedding $(\mathbf{F})$ from a pair of feature maps ($\mathbf{F}_G$, $\mathbf{F}_L$) produced by local and global branches, we suggest combining them by attention-based fusion \cite{hori2017attention}. An intuitive approach is to pool together the information by either average or max operation which has been widely used~\cite{deng2009imagenet,simonyan2014very,he2016deep}. However, considering the complementary nature of local and global attributes, it is more reasonable to adaptively combine the information contained in these two branches. 

As illustrated in Fig.~\ref{fig:moti}, for points ($\mathbf{p}_1$, $\mathbf{p}_2$) located on a plane, local geometric attributes are often ambiguous because of the same relative distances and normals. In this case, the spatial characteristics of ($\mathbf{p}_1$, $\mathbf{p}_2$) described in the global branch are preferred over the local counterpart. By contrast, for the set of points $\mathbf{P}_c$ located around a corner, the local geometric context is distinctive enough and therefore expected to play a more significant role. Inspired by this observation, we have designed a multi-layer perception (MLP) module for attention-based fusion that adaptively integrates two-branch features as follows. 
\begin{figure}[!h]
	\centering
	\includegraphics[width=0.6\linewidth]{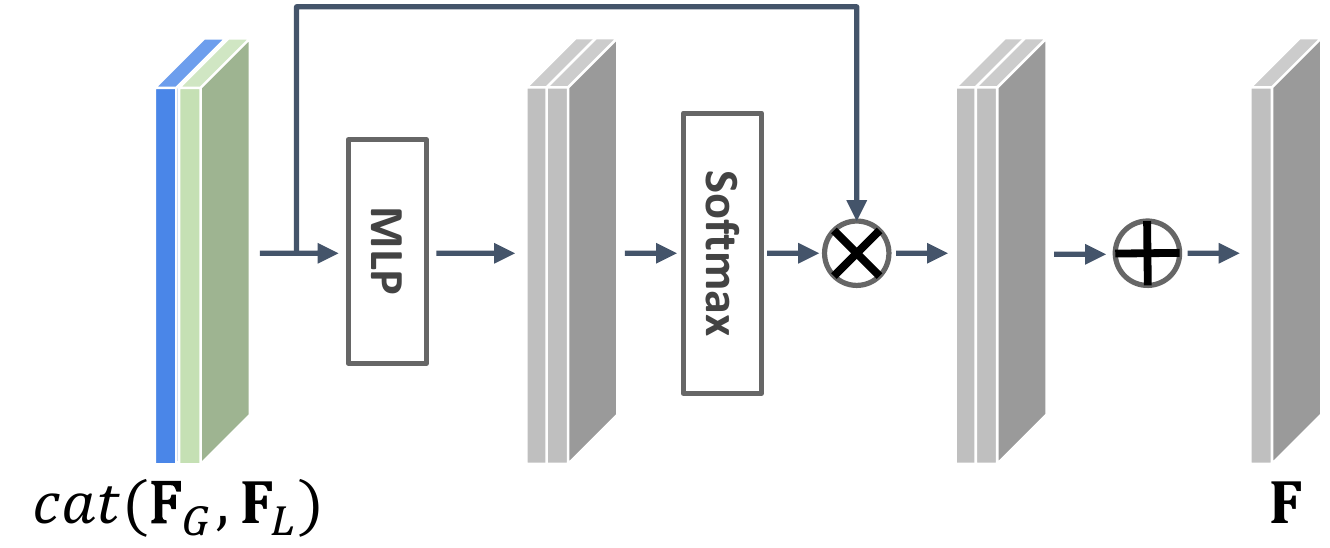}

	\caption{Architecture of the attention-based fusion module.}
	\label{fig:att}
\end{figure}

As shown in Fig.~\ref{fig:att}, $\mathbf{F}_G$ and $\mathbf{F}_L$ are first concatenated and embedded by MLP. Second, a softmax layer is used to estimate the response weights by ($\mathbf{w}_G$ as an example)
\begin{equation}\label{eq:LRF7}
\mathbf{w}_G^{i} = \frac{e^{\mathbf{f}_G^{i}}}{e^{\mathbf{f}_G^{i}} + e^{\mathbf{f}_L^{i}}},
\end{equation}
where $\mathbf{w}_G^{i}$ is the global-branch weight of $\mathbf{p}_i$ and $(\mathbf{f}_G^{i}, \mathbf{f}_L^{i})$ denote embedded features of $\mathbf{p}_i$. Third, the attention-based fusion result $\mathbf{F}$ is generated by 
\begin{equation}\label{eq:LRF8}
\mathbf{F} = \mathbf{w}_{G}\mathbf{F}_G + \mathbf{w}_{L}\mathbf{F}_L.
\end{equation}
Another plausible strategy is immediately concatenating two-branch features and fusing them by MLPs. We empirically suggest that our attention fusion is superior over such strategy as well as the pooling processing (Please refer to Table~\ref{tab:att}).

\subsection{Rotation-Invariant Analysis}
\label{sec:3.5}

\begin{figure}[!h]
	\centering
	\subfigure[Raw points]
	{\includegraphics[width=0.3\linewidth]{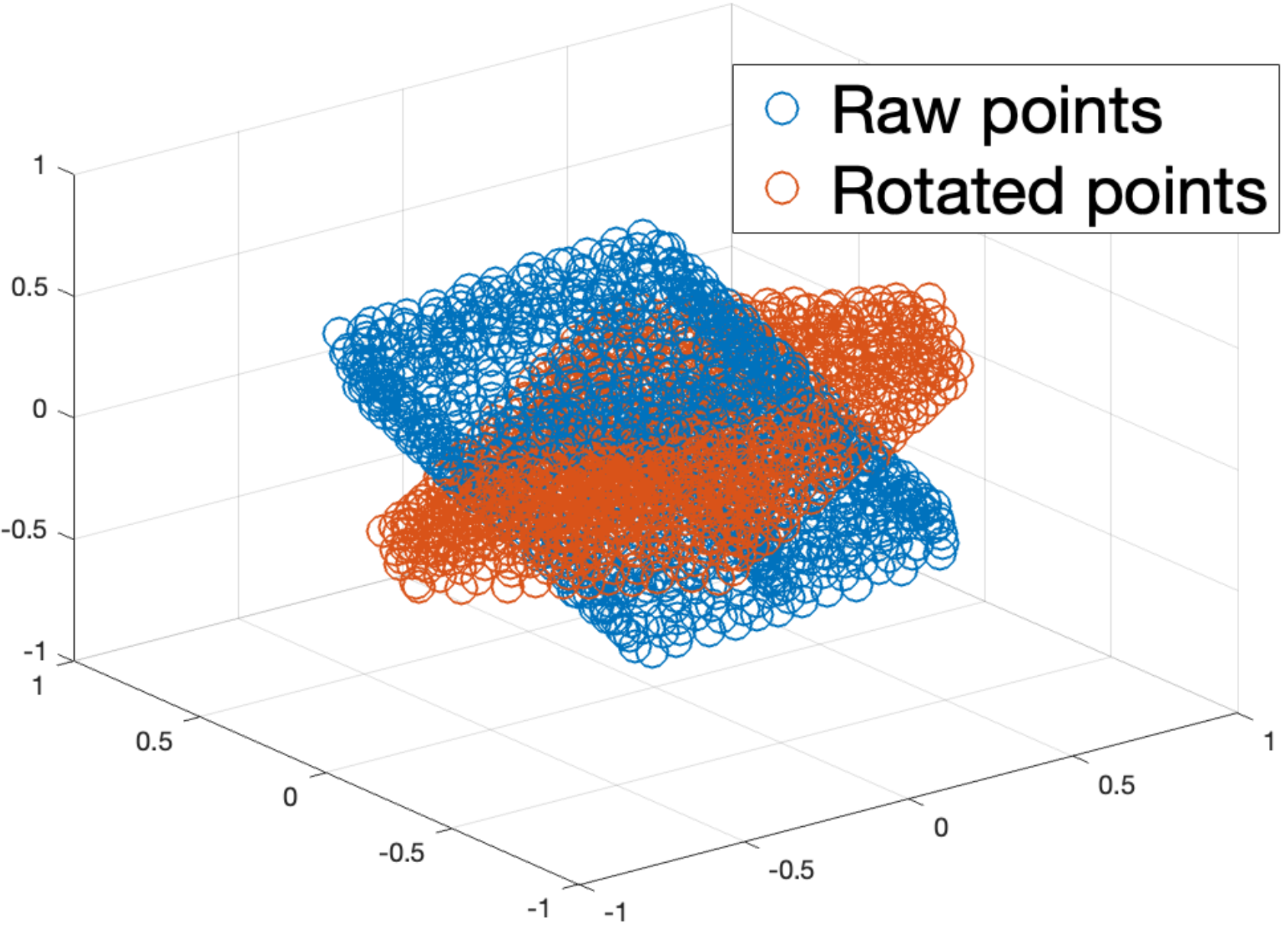}}
	\subfigure[Rotation-invariant representations]
	{\includegraphics[width=0.65\linewidth]{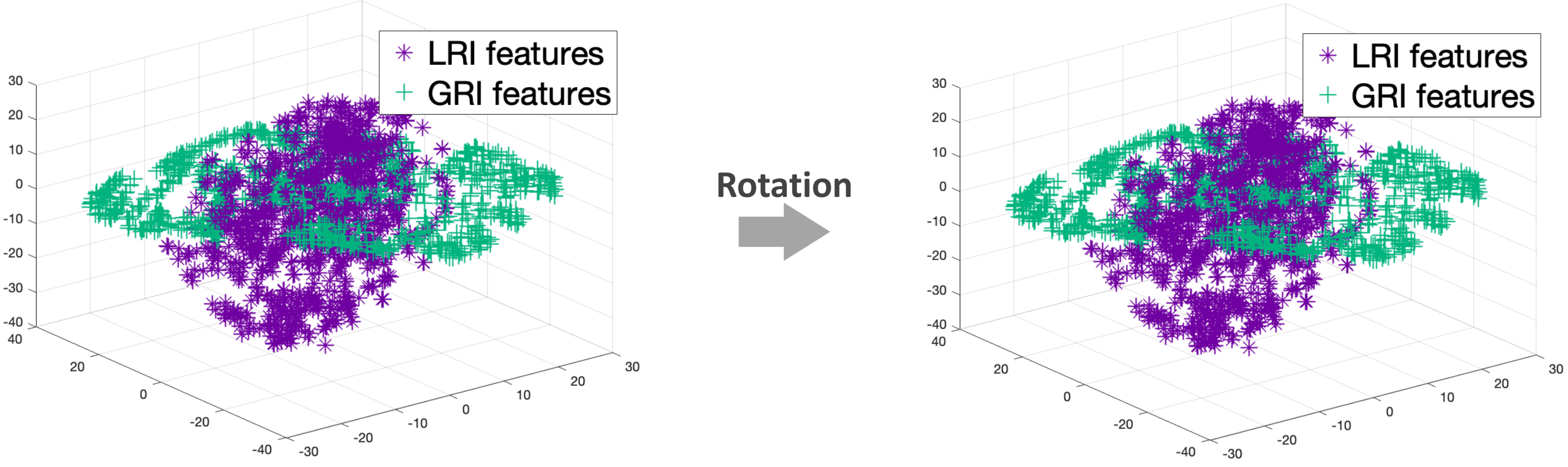}\label{fig_c}}
	\caption{Comparison of the robustness against rotations between (a) raw points and (b) presented rotation-invariant representations. RI representations are visualized in 3D space by $t$-$sne$~\cite{maaten2008visualizing}. Under two different orientations, the raw coordinates of 3D points are significantly changed; while our global and local representations are invariant.}
	\label{fig:rotation}
\end{figure}  
As demonstrated in Fig.~\ref{fig:rotation}, we visualize the extracted global and local representations in the 3D space using $t$-$SNE$~\cite{maaten2008visualizing}. It is straightforward to observe that raw point locations in Fig.~\ref{fig:rotation} (a) are sensitive to orientation changes; while the projected locations of our representations in Fig.~\ref{fig:rotation} (b) are immune to the challenge of rotations. A rigorous justification about the RI properties of distance, angle and SVD transformation is given below.
\\

\noindent\textbf{Distance.  } Assuming $(d, d^{'})$ is the $L_2$ norm of $(\mathbf{p}, \mathbf{p}^{'})$, where $\mathbf{p}^{'}=\mathbf{p}\mathbf{R}$ ($\mathbf{p}\in\mathbb{R}^{1\times3}$), the invariance against rotation is able to be proved as
\begin{equation}\label{eq:LRF9}
{d^{'}=\left\|{\mathbf{p}\mathbf{R}}\right\|=\mathbf{p}\mathbf{R}\mathbf{R}^{T}\mathbf{p}^{T}=d}.
\end{equation}
\\

\noindent\textbf{Angle.  } Supposing $(\theta_{ij}, \theta_{ij}^{'})$ are the angles between $(\mathbf{p}_i, \mathbf{p}_j)$ and $(\mathbf{p}_i^{'}, \mathbf{p}_j^{'})$, the equivalence is formulated as
\begin{equation}\label{eq:LRF10}
{cos(\theta_{ij}^{'}) = \frac {\mathbf{p}_i^{'}\mathbf{p}_j^{'T}}{\left\|\mathbf{p}_i^{'}\right\|\left\|\mathbf{p}_j^{'}\right\|} = \frac {\mathbf{p}_i\mathbf{R}\mathbf{R}^{T}\mathbf{p}_j}{{\left\|\mathbf{p}_i\right\|}\left\|\mathbf{p}_j\right\|} = cos(\theta_{ij})}.
\end{equation}
\\

\noindent\textbf{Singular Value Decomposition.  } We define two point clouds as $\mathbf{P}$ and $\mathbf{P}^{'}$ ($\mathbf{P}, \mathbf{P}^{'}\in\mathbb{R}^{N\times3}$) with $\mathbf{P}^{'} = \mathbf{P}\mathbf{R}$. Singular value decomposition is respectively performed as 
\begin{equation}\label{eq:LRF11}
\mathbf{U}\Sigma\mathbf{V}^{T} = \mathbf{P},
\end{equation}
\begin{equation}\label{eq:LRF12}
\mathbf{U}^{'}\Sigma\mathbf{V}^{'T} = \mathbf{P}^{'},
\end{equation}
where $\mathbf{U}$ and $\mathbf{U}^{'}$ are the eigenvector matrices of $\mathbf{P}\mathbf{P}^{T}$ and $\mathbf{P}^{'}\mathbf{P}^{'T}$, respectively. $\mathbf{U}=\mathbf{U}^{'}$ follows from the symmetry of $\mathbf{P}\mathbf{P}^{T}=\mathbf{P}^{'}\mathbf{P}^{'T}$. The relationship between $\mathbf{V}$ and $\mathbf{V}^{'}$ is able to be derived as $\mathbf{V}^{'}=\mathbf{R}^{T}\mathbf{V}$. The invariance of point locations transformed by $\mathbf{V}$ is then given by

\begin{equation}\label{eq:LRF13}
\mathbf{P}^{'}\mathbf{V}^{'} = \mathbf{P}\mathbf{R}\mathbf{R}^{T}\mathbf{V} = \mathbf{P}\mathbf{V},
\end{equation}

\section{Experimental Results}
\label{sec:exp}
In this section, we report our experimental results on three popular datasets  -i.e., ModelNet40~\cite{wu20153d} (synthetic shape classification), ScanObjectNN~\cite{uy2019revisiting} (real world shape classification), and ShapeNet~\cite{yi2016scalable} (part segmentation). Ablation studies are also included to better illustrate the contribution from each component in our network design.
\subsection{Implementation Details}
For local graph generation, we have used $k$-Nearest-Neighbor (kNN) search to find $32$ neighbors for each central point. In global branch, we down sample the original model to a minimum of $32$ points via farthest point sampling \cite{moenning2003fast}, and an asymmetric edge function is also employed after the transformation as suggested in~\cite{wang2019dynamic}. For further feature extraction, a series of MLPs with increasing dimensions $(64, 128, 512, 1024)$ are employed. Each MLP is followed by Batch Normalization~\cite{ioffe2015batch} and LeakyReLU~\cite{glorot2011deep}. We use three fully connected layers $(512, 256, N_{cls})$ to predict classification results, and three layers of MLPs $(512, 256, N_{seg})$ to generate segmentation results, where $N_{cls}$ and $N_{seg}$ denote the number of candidate labels in classification and segmentation, respectively. The network has been trained for 300 epochs on a NVIDIA TITAN XP GPU using Pytorch with SGD optimizer, learning rate 0.001, batch size 32, following the configuration in~\cite{wang2019dynamic}.
\subsection{Experimental Setup}
For a fair comparison, we divide previous methods into two categories, -i.e., rotation-sensitive and rotation-robust. The experiments are organized into three different conditions, -i.e., raw training data and testing data ($z/z$), raw training data and 3D rotation-augmented testing data ($z/SO3$), and 3D rotation-augmented training data and testing data ($SO3/SO3$). Note that $SO3$ means rotating raw point clouds along three axes, with the aim of taking rotation challenges into account, instead of data augmentation.
\subsection{Synthetic Shape Classification}
We evaluate our method on ModelNet40 which has been extensively used for synthetic shape classification~\cite{li2018so,jiang2018pointsift}. ModelNet40 includes $12311$ CAD models from $40$ categories that are split into $9843$ for training and $2468$ for testing. We randomly sample $1024$ points from each model. These points are then centralized and normalized into an unit sphere. 

Table~\ref{tab:modelnet} includes the experimental results for various experimental settings (two categories and three conditions). First, In the case of $z/z$, our method (LGR-Net) convincingly surpasses all other rotation-robust methods. When compared with Spherical-CNN~\cite{esteves2018learning} and $a^{3}$S-CNN~\cite{liu2018deep} where mesh reconstruction is required, our method achieves superior performance even though we only use raw points as input, which verifies our framework is more effective than spherical solutions. Compared with ClusterNet~\cite{chen2019clusternet}, Riconv~\cite{zhang2019rotation}, which also try to resolve the RI problem, our method still achieves better performance. It confirms that the presented local-global representation (LGR) is more effective. Second, in the situations of $z/SO3$ and $SO3/SO3$, the classification results of LGR-Net are close, outperforming other competitors by a large margin. By contrast, the results of rotation-sensitive algorithms considerably degrade. KPConv~\cite{thomas2019kpconv} that achieves outstanding performance with well-aligned data is vulnerable in $z/SO3$ and $SO3/SO3$. Specifically, it achieves a low accuracy ($18.0\%$) in $z/SO3$ and its performance is still unsatisfying ($87.4\%$) in $SO3/SO3$, even though the training data are augmented by 3D rotations. Moreover, We suggest~\cite{qi2017pointnet,qi2017pointnet++,li2018pointcnn,wang2019dynamic,thomas2019kpconv} are still vulnerable against rotations when points and normals are immediately concatenated as input, because the normal direction is sensitive to rotation. PointNet ($xyz + normal$) is evaluated as an example: the accuracy is $15.9\%$ in $z/
SO3$, which does not look great when compared with PointNet ($xyz$) ($16.4\%$).

\begin{figure}[!h]
	\centering
	\includegraphics[width=0.9\linewidth]{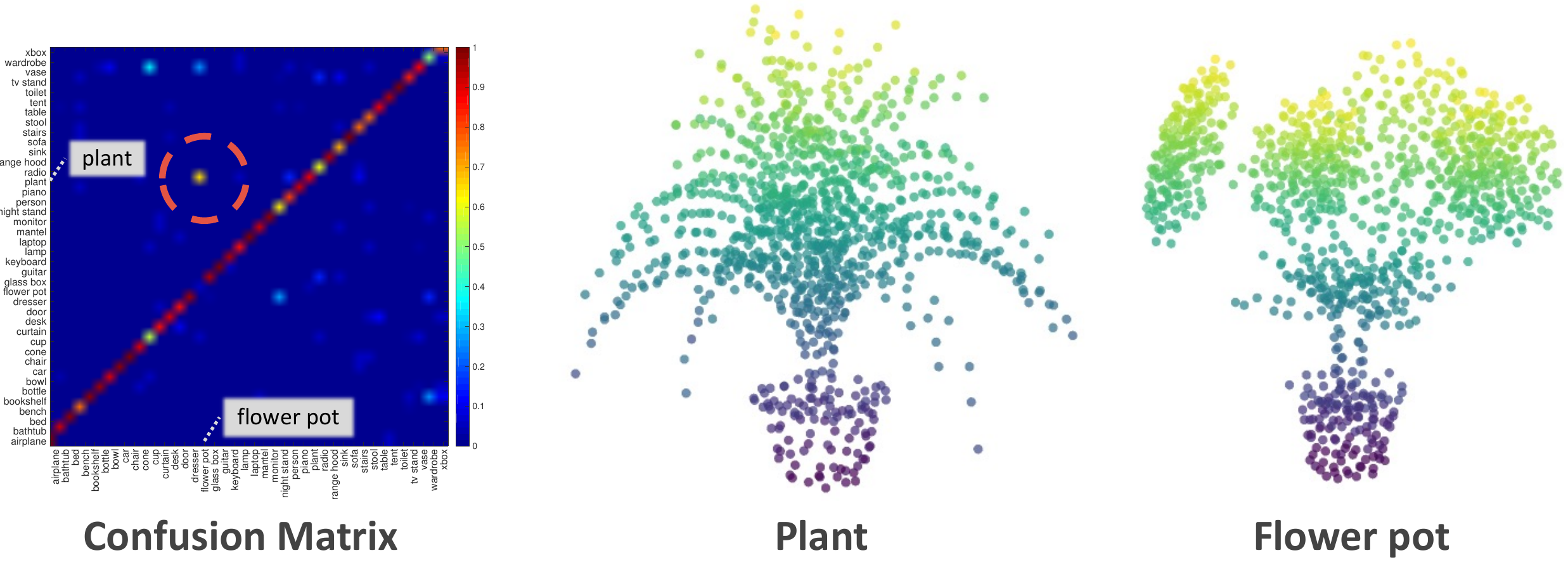}
	\caption{The overlooked intrinsic ambiguity in ModelNet40. The confusion matrix is calculated from the classification result of $40$ candidate categories. The flower pot and plant are the most confusing categories (as verified by the confusion matrix). Both of them include similar plants and pots, which cannot be reliably classified even by human beings. }
	\label{fig:ambi}
\end{figure}

In order to gain a deeper insight into the proposed method, we have calculated the confusion matrix as shown in Fig.~\ref{fig:ambi}. An interesting discovery is that ModelNet40 contains intrinsic ambiguity that has been overlooked in previous studies to the best of our knowledge. More specifically, as illustrated in Fig.~\ref{fig:ambi}, the most two confusing categories are {\em flower pot} and {\em plant}. The exemplar models belonging to these categories are provided, where both two models include similar plants and pots. Even for human observers, these two categories are difficult to distinguish; so it is reasonable for machine-based classification approaches to be confused by such intrinsic ambiguity. 

\begin{table*}[!h]
    \small
	\begin{center}
		\begin{tabular}{lllll}
		    \hline
		    Methods & input  & z/z(\%)& z/SO3(\%) & SO3/SO3(\%) \\
			\hline
			\multicolumn{5}{c}{Rotation-sensitive methods} \\
			\hline
			VoxNet~\cite{maturana2015voxnet} & volume & 83.0 & - & 73.0  \\
			Subvolume~\cite{qi2016volumetric} & volume  & 89.5 & 45.5 & 85.0 \\
	    	MVCNN~\cite{su2015multi} & image & 90.2 & \textbf{81.5} & 86.0 \\
			PointNet~\cite{qi2017pointnet} ($xyz$) & point & 89.2 & 16.4 & 75.5 \\
			PointNet~\cite{qi2017pointnet} ($xyz+normal$) & point & 89.0 & 15.9 & 86.6 \\
			PointNet++~\cite{qi2017pointnet++} & point & 91.8 & 18.4 & 77.4 \\
			PointCNN~\cite{li2018pointcnn} & point & 91.3 & 41.2 & 84.5 \\
			DGCNN~\cite{wang2019dynamic} & point & 92.2 & 20.6 & 81.1 \\
			PointConv~\cite{wu2019pointconv} ($xyz+normal$) & point & 92.5 & 11.7 & 85.9 \\
			KPConv~\cite{thomas2019kpconv} & point & \textbf{92.7} & 18.0 & \textbf{87.4} \\
			\hline
			\multicolumn{5}{c}{Rotation-robust methods} \\
			\hline
			Spherical-CNN~\cite{esteves2018learning} & mesh & 88.9 & 76.7 & 86.9 \\
			$a^{3}$S-CNN~\cite{liu2018deep} & mesh & 89.6 & 87.9 & 88.7 \\
			ClusterNet~\cite{chen2019clusternet} & point & 87.1 & 87.1 & 87.1 \\
			Riconv~\cite{zhang2019rotation} & point & 86.5 & 86.4 & 86.4 \\
		    \textbf{LGR-Net} & point & \textbf{90.9} & \textbf{90.9} & \textbf{91.1} \\
			\hline 
			\end{tabular}
	\end{center}
	\caption{Synthetic shape classification results on ModelNet40. The evaluated approaches are divided into the rotation-sensitive method and rotation-robust method. The experiments are performed in three situations based on different combinations of training data and testing data. $z$ and $SO3$ are respectively represent raw data and 3D rotation-augmented data. The metric is classification accuracy and the best result is rendered in bold.}
	\label{tab:modelnet}
\end{table*}

\subsection{Real World Shape Classification}

\begin{figure}[!h]
	\centering
	\includegraphics[width=0.8\linewidth]{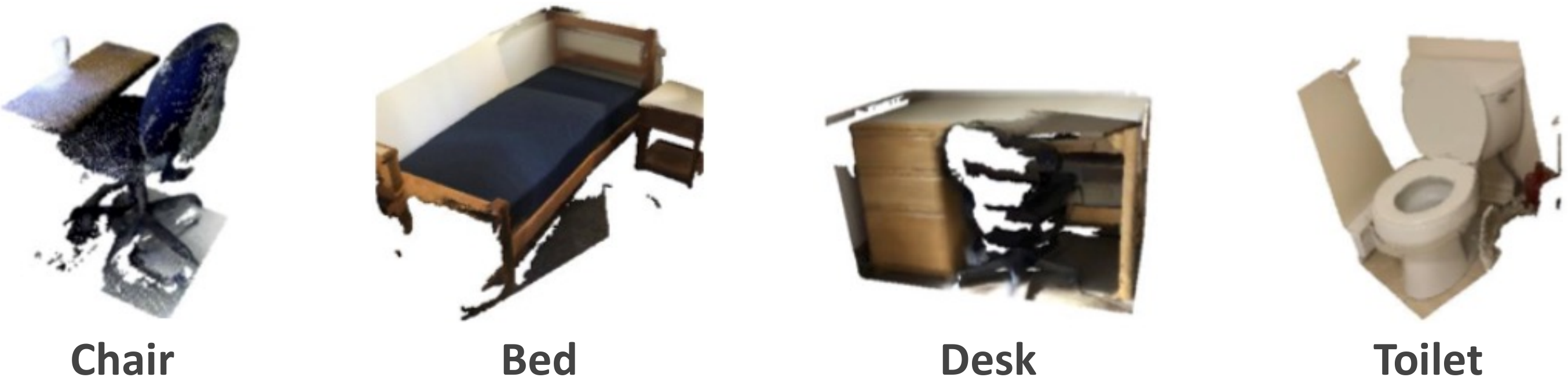}
	\caption{Object examples in ScanObjectNN, where some typical nuisances, i.e., missing data, occlusions, and non-uniform density are included.}
	\label{fig:examples}
\end{figure}

Considering that the objects in ModelNet40 are man-made CAD models (well-aligned and noise-free), there is a significant gap between synthetic data and real-world data. Real-world point clouds tend to include a variety of nuisances -e.g., missing data, occlusions, and non-uniform density. In order to evaluate rotation invariance in such conditions, we have conducted experiments on ScanObjectNN~\cite{uy2019revisiting} which consists of real-world indoor scenes. This dataset includes $2902$ objects which are classified into $15$ categories. Some examples taken from this dataset are shown in Fig.~\ref{fig:examples}.

We have conducted the experiments on both the easiest subset \emph{OBJ\_BG} (without rotation, translation, and scaling) and the hardest subset \emph{PB\_T50\_RS} (contains $50\%$ bounding box translation, rotation around the gravity axis, and random scaling). Note that Spherical-CNN and $a^{3}$S-CNN are not evaluated on this dataset because the required mesh data are unavailable; ClusterNet is unable to be retrained because of the unreleased code. The comparative results in terms of classification accuracy are shown in Table~\ref{tab:scan}. In the cases of $z/SO3$ and $SO3/SO3$, our method achieves the best performance, which indicates that LGR-Net is not only invariant to rotation, but also robust to typical nuisances, i.e., missing data, occlusions, and non-uniform density. 
\begin{table*}[!h]
    \small
	\begin{center}
		\begin{tabular}{l|llllll}
		    \hline
		    \multirow{2}*{Methods} & \multicolumn{3}{c}{OBJ\_BG (\%)} & \multicolumn{3}{c}{PB\_T50\_RS (\%)} \\
		    \cline{2-7}
		       ~    & z/z & z/SO3 & SO3/SO3 & z/z & z/SO3 & SO3/SO3 \\
			\hline
			\multicolumn{7}{c}{Rotation-sensitive methods} \\
			\hline
            PointNet~\cite{qi2017pointnet} & 73.3 & 16.7 & 54.7 & 68.2 & \textbf{17.1} & 42.2 \\
            PointNet++~\cite{qi2017pointnet++} & 82.3 & 15.0 & 47.4 & 77.9 & 15.8 & 60.1 \\
            PointCNN~\cite{li2018pointcnn} & \textbf{86.1} & 14.6 & 63.7 & \textbf{78.5} & 14.9 & 51.8 \\
            DGCNN~\cite{wang2019dynamic} & 82.8 & \textbf{17.7} & \textbf{71.8} & 78.1 & 16.1 & \textbf{63.4} \\
			\hline 
			\multicolumn{7}{c}{Rotation-robust methods} \\
			\hline
			Riconv~\cite{zhang2019rotation} & 78.4 & 78.4 & 78.1 & 67.9 & 67.9 & 68.3 \\
			\textbf{LGR-Net} & \textbf{81.2} & \textbf{81.2} & \textbf{81.4} & \textbf{72.7} & \textbf{72.7} & \textbf{72.9} \\
			\hline 
			\end{tabular}
	\end{center}
	\caption{Real world shape classification results on ScanObjectNN. Two parts, -i.e., \emph{OBJ\_BG} and \emph{PB\_T50\_RS} are considered. \emph{OBJ\_BG} contains objects and backgrounds without rotation, translation, and scaling; while \emph{PB\_T50\_RS} takes into account of $50\%$ bounding box translation, rotation around the gravity axis, and random scaling. The performance is measured by classification accuracy.}
	\label{tab:scan}
\end{table*}
\subsection{Part Segmentation}
Given a point cloud model, the objective of segmentation is to accurately predict per-point labels. When compared with the shape classification, segmentation is more challenging because it involves the discrimination of fine-detailed structures. We have extended our experiments on ShapeNet~\cite{yi2016scalable} - a widely used dataset for evaluating part segmentation. We have considered a subset of ShapeNet including $16881$ 3D models, $16$ kinds of objects, and $50$ part categories.  The average category mIoU (Cat. mIoU)~\cite{shen2018mining} is utilized to compare the segmentation performance.
\begin{table*}[!h]
    \footnotesize
    \begin{center}
		\begin{tabular}{c|c|c|c|c|c|c|c|c|c|c|c|c|c|c|c|c}
		    \hline
		     & aero & bag & cap & car & chair & earph. & guitar & knife & lamp & laptop & motor & mug & pistol & rocket & skate & table \\
			\hline
			\#shapes & 2690 & 76 & 55 & 898 & 3758 & 69 & 787 & 392 & 1547 & 451 & 202 & 184 & 283 & 66 & 152 & 5271  \\
			\hline   
		    & \multicolumn{16}{c}{$z/SO3 (\%)$} \\
			\hline
			PointNet~\cite{qi2017pointnet} & 40.4 & 48.1 & 46.3 & 24.5 & 45.1 & 39.4 & 29.2 & 42.6 & 52.7 & 36.7 & 21.2 & 55.0 & 29.7 & 26.6 & 32.1 & 35.8  \\
			PointNet++~\cite{qi2017pointnet++} & 51.3 & 66.0 & 50.8 & 25.2 & 66.7 & 27.7 & 29.7 & 65.6 & 59.7 & 70.1 & 17.2 & 67.3 & 49.9 & 23.4 & 43.8 & 57.6 \\
			PointCNN~\cite{li2018pointcnn} & 21.8 & 52.0 & 52.1 & 23.6 & 29.4 & 18.2 & 40.7 & 36.9 & 51.1 & 33.1 & 18.9 & 48.0 & 23.0 & 27.7 & 38.6 & 39.9 \\
			DGCNN~\cite{wang2019dynamic} & 37.0 & 50.2 & 38.5 & 24.1 & 43.9 & 32.3 & 23.7 & 48.6 & 54.8 & 28.7 & 17.8 & 74.4 & 25.2 & 24.1 & 43.1 & 32.3 \\
			\hline
			Riconv~\cite{zhang2019rotation} & \underline{80.6} & \underline{80.0} & \underline{70.8} & \underline{68.8} & \underline{86.8} & \underline{70.3} & \underline{87.3} & \textbf{84.7} & \underline{77.8} & 80.6 & \underline{57.4} & \underline{91.2} & \underline{71.5} & \underline{52.3} & \underline{66.5} & \underline{78.4} \\
			\textbf{LGR-Net} & \textbf{81.5} & \textbf{80.5} & \textbf{81.4} & \textbf{75.5} & \textbf{87.4} & \textbf{72.6} & \textbf{88.7} & \underline{83.4} & \textbf{83.1} & \textbf{86.8} & \textbf{66.2} & \textbf{92.9} & \textbf{76.8} & \textbf{62.9} & \textbf{80.0} & \textbf{80.0}\\
			\hline
			& \multicolumn{16}{c}{$SO3/SO3 (\%)$} \\
		    \hline
			PointNet~\cite{qi2017pointnet} & \underline{81.6} & 68.7 & 74.0 & 70.3 & \underline{87.6} & 68.5 & \underline{88.9} & 80.0 & 74.9 & 83.6 & 56.5 & 77.6 & 75.2 & \underline{53.9} & \underline{69.4} & 79.9 \\
            PointNet++~\cite{qi2017pointnet++} & 79.5 & 71.6 & \textbf{87.7} & \underline{70.7} & \textbf{88.8} & 64.9 & 88.8 & 78.1 & 79.2 & \textbf{94.9} & 54.3 & \underline{92.0} & 76.4 & 50.3 & 68.4 & \underline{81.0} \\
            PointCNN~\cite{li2018pointcnn} & 78.0 & \underline{80.1} & 78.2 & 68.2 & 81.2 & 70.2 & 82.0 & 70.6 & 68.9 & 80.8 & 48.6 & 77.3 & 63.2 & 50.6 & 63.2 & \textbf{82.0} \\
            DGCNN~\cite{wang2019dynamic} & 77.7 & 71.8 & 77.7 & 55.2 & 87.3 & 68.7 & 88.7 & \underline{85.5} & \underline{81.8} & 81.3 & 36.2 & 86.0 & \underline{77.3} & 51.6 & 65.3 & 80.2 \\
			\hline
			Riconv~\cite{zhang2019rotation} & 80.6 & \textbf{80.2} & 70.7 & 68.8 & 86.8 & \underline{70.4} & 87.2 & 84.3 & 78.0 & 80.1 & \underline{57.3} &  91.2 & 71.3 & 52.1 & 66.6 & 78.5 \\
			\textbf{LGR-Net} & \textbf{81.7} & 78.1 & \underline{82.5} & \textbf{75.1} & \underline{87.6} & \textbf{74.5} & \textbf{89.4} & \textbf{86.1} & \textbf{83.0} & 86.4 & \textbf{65.3} & \textbf{92.6} & 75.2 & \textbf{64.1} & \textbf{79.8} & 80.5   \\
			\hline
			\end{tabular}
	\end{center}
	\caption{Specific per-class average mIoU in the cases of $z/SO3$ and $SO3/SO3$ (\textbf{bold} - the best, \underline{underline} - the second best). }
	\label{tab:seg}
\end{table*}

\begin{table}[!h]
    \small
	\begin{center}
		\begin{tabular}{llll}
		    \hline
		    Methods & z/z (\%) & z/SO3 (\%) & SO3/SO3 (\%)\\
			\hline
			\multicolumn{4}{c}{Rotation-sensitive methods} \\
			\hline
			PointNet~\cite{qi2017pointnet} & 80.4 & 37.8 & 74.4 \\
			PointNet++~\cite{qi2017pointnet++} & 81.9 & \textbf{48.2} & \textbf{76.7} \\
			PointCNN~\cite{li2018pointcnn} & \textbf{84.6} & 34.7 & 71.4 \\
			DGCNN~\cite{wang2019dynamic} & 82.3 & 37.4 & 73.3 \\
			\hline
			\multicolumn{4}{c}{Rotation-robust methods} \\
			\hline
			Riconv~\cite{zhang2019rotation} & 74.6 & 75.3 & 75.5 \\
			\textbf{LGR-Net} & \textbf{80.0} & \textbf{80.0} & \textbf{80.1} \\
			\hline
			\end{tabular}
	\end{center}
	\caption{Overall part segmentation results on ShapeNet. The metric is overall average category mIoU (Cat. mIoU) estimated by averaging the results over $16$ categories.}
	\label{tab:seg2}
\end{table}

Specific results in two different cases are shown in Table~\ref{tab:seg}. In the case of $z/SO3$, our LGR-Net dramatically surpasses the previous rotation-robust method Riconv (ours behaves better in 15 categories out of 16 ones in total). In $SO3/SO3$, LGR-Net achieves the most consistent performance, significantly exceeding other methods (for 12 out of 16 categories, ours achieves the best or the second best result). The overall comparison results are reported in Table~\ref{tab:seg2}, which clearly justify the superiority of LGR-Net.

\begin{figure}[!t]
	\centering
	\includegraphics[width=1.0\linewidth]{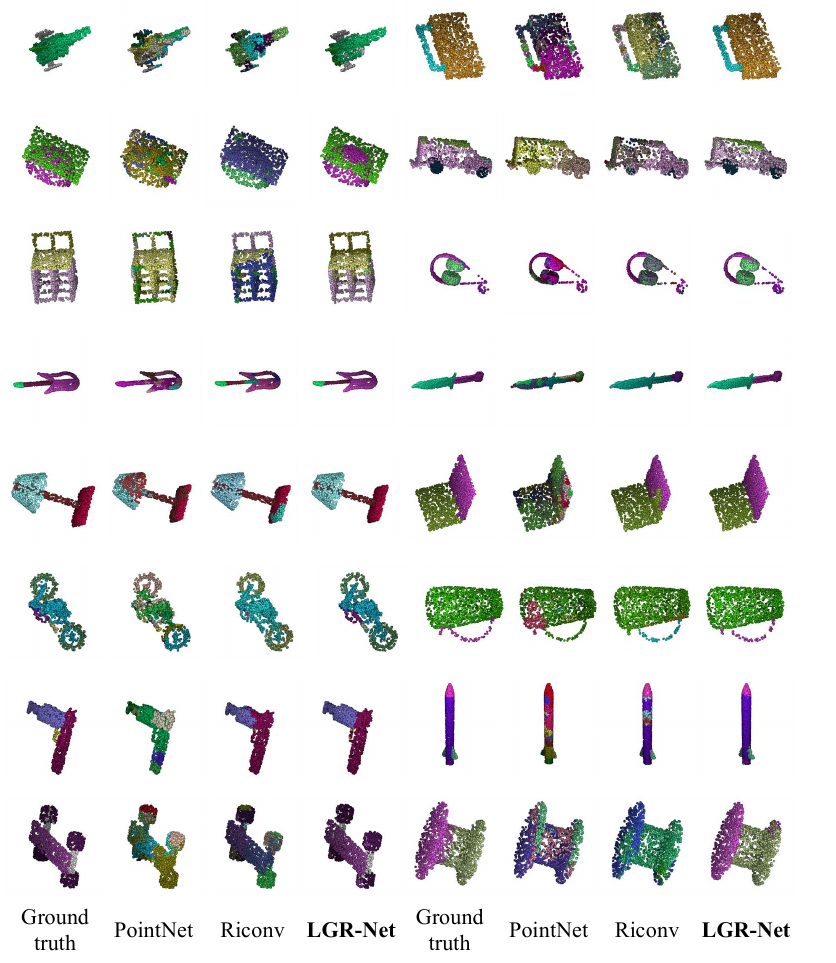}
	\caption{Visualization results for part segmentation. The networks are trained and tested on ShapeNet, and different parts are represented by different colors. The training data are rotation-free, while the testing data are transformed by the specific 3D rotation, -i.e., $(45^{\circ}, 45^{\circ}, 45^{\circ})$.}
	\label{fig:v}
\end{figure}

Some representative visualization results for part segmentation on ShapeNet are shown in Fig.~\ref{fig:v}. The training data are rotation-free, while the testing data are transformed by the specific 3D rotation, -i.e., $(45^{\circ}, 45^{\circ}, 45^{\circ})$. It can be easily observed that our approach (LGR-Net) significantly outperforms the competing methods (PointNet \cite{qi2017pointnet} and Riconv \cite{zhang2019rotation}). Our part segmentation results are visually closer to the ground-truth in all 16 different cases.

\subsection{3D Scene Segmentation}
To shed more light on the effectiveness of our method in real-world applications, we perform experiments on S3DIS~\cite{armeni20163d}, which has been widely used as a benchmark of 3D Scene segmentation. Specifically, S3DIS includes 273 million points in total, which are sampled from six large-scale indoor areas and divided into 13 classes. We employ Area-5 for testing and use the other five areas for training, following the settings in~\cite{tchapmi2017segcloud, thomas2019kpconv}. Since normals are not provided on S3DIS, we take the distances coupled with RGB features as input in our local branch. 

As reported in Table~\ref{tab:s3dis}, the rotation-sensitive methods are vulnerable against rotations on S3DIS. The mIoU of DGCNN drops by 44.8\% when rotations are taken into account, i.e., from 48.4\% in $z/z$ to 3.6\% in $z/SO3$. The performance is still limited in $SO3/SO3$, i.e., $34.3\%$, even though the training data are 3D rotation-augmented. By contrast, our LGR-Net achieves superior results when facing the challenge of rotations, outperforming the previous rotation-robust method, i.e., Riconv, by 21.4\%. Moreover, since our global rotation-invariant features are extracted based on SVD on down-sampled 32 points (default setting), one may concern if the presented global RI features are effective on large-scale point clouds. To address this issue, we perform an ablation study on S3DIS, which removes global branch from the proposed network and only takes the local RI features as input. Empirically, we found the removing of global branch results in 7.0\% mIoU decrease, i.e., from 43.4\% to 36.4\%. It demonstrates that the global branch still plays an important role in 3D scene segmentation. 

\begin{table}[!h]
    \small
	\begin{center}
		\begin{tabular}{llll}
		    \hline
		    Methods & z/z (\%) & z/SO3 (\%) & SO3/SO3 (\%)\\
			\hline
			\multicolumn{4}{c}{Rotation-sensitive methods} \\
			\hline
			PointNet~\cite{qi2017pointnet} & 41.1 & 4.1 & 29.3 \\
			DGCNN~\cite{wang2019dynamic} & \bf{48.4} & 3.6 & 34.3 \\
			\hline
			\multicolumn{4}{c}{Rotation-robust methods} \\
			\hline
			Riconv~\cite{zhang2019rotation} & 22.0 & 22.0 & 22.0 \\
			\textbf{LGR-Net} & \bf{43.4} & \bf{43.4} & \bf{43.4} \\
			\hline
			\end{tabular}
	\end{center}
	\caption{3D scene segmentation results (mIoU) on S3DIS Area-5~\cite{armeni20163d}.}
	\label{tab:s3dis}

\end{table}

\subsection{Robustness against sampling rates and strategies}
Since our global RI representation is generated on three orthogonal axes which are estimated by SVD on a down-sampling structure, the robustness of SVD against different down-sampling rates and strategies is critical for our method. To address this issue, we evaluate LGR-Net on $OBJ\_BG$ of ScanObjectNN, employing two different down-sampling strategies and three down-sampling rates. The results are shown in Table~\ref{tab:rob}.
\begin{table}[!h]
    \small
	\begin{center}
		\begin{tabular}{l|l|l|l}
		    \hline
		     \multirow{2}*{Sampling strategy} & \multicolumn{3}{c}{\# points} \\
		    \cline{2-4}
		    ~ & 8 & 16 & 32 \\
			\hline			Random sampling (\%) & 79.0 & 79.0 & 80.9 \\
			Farthest point sampling (\%) & 78.0 & 80.7 & 81.2  \\
			\hline
			\end{tabular}
	\end{center}
	\caption{Analysis of robustness of SVD against sampling rates and strategies on $OBJ\_BG$. We sample 32 points at most because of the computational cost.}
	\label{tab:rob}
\end{table}
Under different down-sampling strategies and down-sampling rates, our method still outperforms RiConv (78.4\%) in most cases, which justifies that our global RI representation is robust to sampling strategies and sampling rates.
\subsection{Ablation Studies}
\label{sec:abalation}
Ablation studies are performed to demonstrate the rationality of our network design. Specifically, we separately train the global branch and local branch of the LGR-Net classification network on \emph{OBJ\_BG} of ScanObjectNN. We also train two other versions, replacing attention-based fusion module by an ad-hoc average pooling layer (Avg-Pool) and concatenation-and-convolution processing (Cat-Conv), respectively. As shown in Table~\ref{tab:ablation} and Table~\ref{tab:att}, the results suggest that both branches and attention-based fusion have positive impacts on LGR-Net. 
The two branches contain complementary information -i.e., local geometry and global topology; attention-based fusion is capable of combining them adaptively and making the feature fusion process more reasonable. 
\begin{table}[!h]
    \footnotesize
	\begin{center}
		\begin{tabular}{l|l|l|l}
		    \hline
		    Global Branch & Local Branch & Attention Fusion & Accuracy (\%) \\
			\hline
			\checkmark & $\times$ & $\times$ & 71.6 \\
			$\times$ & \checkmark & $\times$ & 71.3 \\
			\checkmark & \checkmark & $\times$ & 80.6 \\
			\checkmark & \checkmark & \checkmark & \textbf{81.2} \\
			\hline
			\end{tabular}
	\end{center}
	\caption{Ablation studies of LGR-Net on \emph{OBJ\_BG}.}
	\label{tab:ablation}
\end{table}
\begin{table}[!h]
    \small
	\begin{center}
		\begin{tabular}{l|l|l|l}
		    \hline
		    Method & Avg-Pool & Cat-Conv & Attention \\
			\hline
			Accuracy (\%) & 80.6 & 80.2 & 81.2 \\
			\hline
			\end{tabular}
	\end{center}
	\caption{Analysis of fusion methods, where Cat-Conv concatenates two-branch features and then fuses the features by a convolution layer.}
	\label{tab:att}
\end{table}

We have implemented an one-branch version at the early stage of this work, which projects all points into the estimated global coordinate system and then performs local feature extraction (One-Branch Network). 
\begin{table}[!h]
    \small
	\begin{center}
		\begin{tabular}{l|l|l|l}
		    \hline
		    Method & z/z (\%) & z/SO3 (\%) & SO3/SO3 (\%) \\
			\hline
			One-Branch Network & 78.7 & 78.7 & 79.2 \\
			LGR-Net & 81.2 & 81.2 & 81.4 \\
			\hline
			\end{tabular}
	\end{center}
	\caption{Analysis of the two-branch design on \emph{$OBJ\_BG$}. The one-branch version projects all points into the estimated global coordinate system and then performs local feature extraction.}
	\label{tab:one}
\end{table}
This one-branch version has been abandoned because we are motivated by the observation that local and global information are supposed to play complementary roles in different regions. Consequently, it is more reasonable to separately extract local and global information in a two-branch architecture and fuse the feature maps in an adaptive manner. The results of One-Branch Network and LGR-Net on $OBJ\_BG$ of ScanObjectNN are shown in Table~\ref{tab:one}. Compared with One-Branch Network, LGR-Net performs better in all three cases. The experimental results have justified the superiority of our new two-branch design.
\section{Limitations and Discussions}
To analyze limitations of our method, we replace RI representations in the global branch with well-aligned point coordinates. The performance on the rotation-free version of $OBJ\_BG$ significantly increases (around $6\%$). Based on such observation, we discuss two  limitations of our global RI representations as follows.

First, considering that the objects in existing datasets are well-aligned, the poses among different instances from the same category are consistent, which provide an underlying consistency for learning. Although the projected poses in the global branch are invariant to rotation, we find that the orientations of objects in the same category are not quite consistent, increasing the difficulty of learning. A method which is not only invariant to rotation but also can align instances from the same category to a consistent orientation is desired in our further work.

Second, real-world point cloud data are at the mercy of noise contamination and missing data (e.g., due to the occlusion or reflection from specular surfaces). Orientations determined by SVD can be affected by such nuisances. At present, we alleviate this issue by performing SVD on down-sampled skeleton-like structures which exhibit certain robustness against noise and missing data. Nonetheless, how to overcome the influence of such nuisances remains an important direction for improving our work.

Although LGR-Net has above-mentioned limitations, we have achieved a better trade-off between the accuracy/mIoU and rotation invariance when compared with other competing methods as confirmed by experimental results in Sec.~\ref{sec:exp}. 

\section{Conclusion}
We have presented RI representations in terms of local geometry and global topology for point cloud analysis. With RI feature extraction, we integrate the representations into a two-branch network, where an attention-based fusion module is designed to adaptively fuse two-branch features. Both theoretical and empirical proofs for RI are provided. Experimental results have demonstrated the superiority of our network design to other competing approaches. 
In our future works, we expect to study the adaptation of LGR-Net to large-scale datasets (e.g., KITTI) to facilitate other point cloud based vision tasks such as LiDAR SLAM and autonomous driving.
\section*{Acknowledgment}
This work was supported in part by the National Natural Science Foundation of China under Grant 61876211 and by the 111 Project on Computational Intelligence and Intelligent Control under Grant B18024. Xin Li's work is partially supported by the DoJ/NIJ under grant NIJ 2018-75-CX-0032, NSF under grant OAC-1839909, IIS-1951504 and the WV Higher Education Policy Commission Grant (HEPC.dsr.18.5).


\bibliography{mybibfile}
\end{sloppypar}

\end{document}